\documentclass[preprint,authoryear,12pt]{elsarticle}
\usepackage{mathrsfs}
\usepackage{graphics,graphicx,epsfig}
\usepackage{amsmath,mathrsfs,amssymb,amsfonts}
\usepackage{algorithmic}
\usepackage{multirow,bm}
\usepackage{multirow}
\usepackage[ruled]{algorithm2e}
\usepackage{float}
\usepackage{subfloat}
\usepackage{leftidx}
\usepackage{natbib}
\usepackage{amsmath}
\usepackage{subfig}
\usepackage{hyperref}
\usepackage{multirow}
\usepackage{caption}
\usepackage{amsmath,amsthm,amssymb,amsfonts}
\newtheorem{theorem}{Theorem}

\journal{arXiv}

\begin{document}
\begin{frontmatter}

\title{AutoEncoder by Forest}

\author{Ji Feng and Zhi-Hua Zhou\corref{cor1}}

\address{National Key Laboratory for Novel Software Technology\\
             Nanjing University, Nanjing 210023, China}
\cortext[cor1]{Email: zhouzh@lamda.nju.edu.cn}

\begin{abstract}
Auto-encoding is an important task which is typically realized by deep neural networks (DNNs) such as convolutional neural networks (CNN). In this paper, we propose EncoderForest (abbrv. eForest), the first tree ensemble based auto-encoder. We present a procedure for enabling forests to do backward reconstruction by utilizing the equivalent classes defined by decision paths of the trees, and demonstrate its usage in both supervised and unsupervised setting. Experiments show that, compared with DNN autoencoders, eForest is able to obtain lower reconstruction error with fast training speed, while the model itself is reusable and damage-tolerable.
\end{abstract}

\end{frontmatter}
\section{Introduction}

Auto-encoder \citep{vincent2010stacked} is a class of models which aim to map the input to a latent space and map it back to the original space, with low reconstruction error as its objective. Previous approaches for building such device mainly came from the neural network community. For instance, a neural network based auto-encoder usually consists of an encoder and a decoder. The encoder maps the input to a hidden layer and the decoder maps it back to the input space. By concatenating the two parts and setting the reconstruction error as learning objective, back-propagation can be used for training such models. It is widely used for dimensionality reduction \citep{hinton2006fast}, representation learning \citep{bengio2013representation}, as well as some more recent works in generative models such as Variational Auto-encoders \citep{kingma2013auto}.

Ensemble learning \citep{Zhou2012} is a powerful learning paradigm which trains multiple learners and combines to tackle the problem. It is widely used in a broad range of tasks and and demonstrates great performance. Tree ensemble methods, or forests, such as Random Forest \citep{Breiman2001}, for instance, is one of the best off-the-shelf methods for supervised learning \citep{fernandez2014we}. Other successful tree ensembles such as gradient based decision trees (GBDTs), \citep{chen2016xgboost} has also proven its ability during the past decade.
Besides supervised learning, tree ensembles have also achieved great success in other tasks, such as isolation forest \citep{Liu:Ting:Zhou2008} which is an efficient unsupervised method for anomaly detection. Recently, deep model based on forests has also been proposed \citep{Zhou:gcForest17}, and demonstrated competitive performance with DNNs across a broad range of tasks with much fewer hyper-parameters.

In this paper, we present the EncoderForest, (abbrv. eForest), by enabling a tree ensemble to perform forward encoding and backward decoding operations and can be trained in both supervised or unsupervised fashion. Experiments showed the eForest approach has the following advantages:
\begin{itemize}
  \item Accurate: Its experimental reconstruction error is lower than a MLP or CNN based auto-encoders.
  \item Efficient: eForest on a single KNL (many-core CPU) runs even faster than a CNN auto-encoder runs on a Titan-X GPU for training.
  \item Damage-tolerable: The trained model works well even when it is partially damaged.
  \item Reusable: A model trained from one dataset can be directly applied on the other dataset in the same domain.
\end{itemize}

The rest of the paper is organized as follows: first we introduce related works, followed by the proposed eForest model, then experimental results are presented, finally conclusion and future works are discussed.

\section{Related Work}

Auto-encoding an important task for learning association from data, which is one of the key ingredient of deep learning. \citep{Goodfellow:Bengio2016}.
The study of auto-encoding dates back to \citep{bourlard1988auto}, of which the goal is to learning an auto-association relation which can be used to for representation learning.
\citep{bengio2013representation}.  Most of the previous approaches on auto-encoding are neural network based models. For instance, the under-complete auto-encoder, which purpose is to compress data for dimensionality reduction \citep{hinton2006reducing} and efficient coding\citep{liou2008modeling}, sparse auto-encoder gives a sparsity penalty on the on the activation layer \citep{hinton2010modeling}, which is related with sparse coding \citep{willmore2001characterizing}, and denoising auto-encoders \citep{bengio2013generalized} forces the model to learn the mapping from a corrupted input to its noiseless version. Applications ranging from computer vision \citep{masci2011stacked} to natural language processing \citep{mikolov2013distributed} and semantic hashing \citep{salakhutdinov2007restricted} which uses autoencoders in information retrieval tasks. In fact, the concept of deep learning stated with training a stack of auto-encoders in a greedy layer-wised fashion. \citep{hinton2006fast}. Auto-encoding has also been applied in some more recent works such as variational auto-encoder for generative models \citep{kingma2013auto}.

Ensembles of decision trees, or called \textit{forest}, are popularly used in ensemble learning \citep{Zhou2012}. For example, Bagging \citep{Breiman1996} and Boosting \citep{Freund:Schapire:1999} usually take decision trees as component learners. Other famous decision tree ensemble methods including Random Forest \citep{Breiman2001} and GBDT\citep{Friedman2001}; the former is a variant of Bagging, whereas the latter is a variant of Boosting. Some efficient implementations of GBDT, e.g. XGBoost \citep{chen2016xgboost}, has been widely used in industry and various data analytics competitions. In addition to the above tree ensembles constructed in supervised setting, there are unsupervised tree ensembles also proven to be useful in various domains. For example, the iForest \citep{Liu:Ting:Zhou2008} is an unsupervised forest designed for anomaly detection, and its ingredient, completely-random decision trees, have also been applied to tasks such as streaming new class learning \citep{Mu:Ting:Zhou2017}. Note that both supervised and unsupervised forests, i.e. Random Forest and completely-random tree forest, have been simultaneously exploited in the construction of deep forest\citep{Zhou:gcForest17}.

\section{The Proposed Method}

An auto-encoder has two basic functions: encoding and decoding. There is no difficulty for a forest to do encoding, because at least the leaf nodes information can be regarded as a kind of encoding; needless to say, the subsets of nodes or even the branch of paths may be able to offer more information for encoding.

First, we propose the encoding procedure of EncoderForest. Given a trained tree ensemble model of $T$ trees, the forward encoding procedure takes an input data and send this data to each root node of trees in the ensemble, once the data traverse down to the leaf nodes for all trees, the procedure will return a $T$ dimensional vector, where each element $t$ is an integer index of the leaf node in tree $t$.

A more concrete algorithm for forward encoding is shown in Algorithm~\ref{alg:Encoding}.
Notice that this encoding procedure is independent with the particular learning rule on how to split the nodes for trees. For instance, the decision rule can be learned in a supervised setting such as random forest, or can be learned in an unsupervised setting such as completely random trees.

\begin{algorithm}[H]
\caption{Forward Encoding}
\label{alg:Encoding}
\KwIn{A trained forest $F$ with $T$ trees, an input data $x$}
\KwOut{$x_{enc}$}

 $x_{enc}$ = $zeros$[$T$,$1$]         \%  initialize $x'$ \\
\For{$i$ in range($T$)}
{\
     $x_{enc}[i]$ = $Forest.tree[i].encode(x)$ \\
    \% returns leaf index for tree $i$ \\
}
return  $x_{enc}$
\end{algorithm}

On the other hand, however, the decoding function is not that obvious. In fact, forests are generally used for forward prediction, by going from the root of each tree to the leaves, whereas it is unknown how to do backward reconstruction, i.e., inducing the original samples from information obtained at the leaves.

Suppose we are handling a binary classification task, with four attributes. The first and second attributes are numerical ones; the third is a boolean attribute with values {\textit{YES}, \textit{NO}}; the fourth is a triple-valued attribute with values {\textit{RED}, \textit{BLUE}, \textit{GREEN}}.  Given an instance $\bm{x}$, let $x_i$ denotes the value of $\bm{x}$ on the $i$-th attribute.

\begin{figure*}[t]
   \centering
   \includegraphics[scale=0.25]{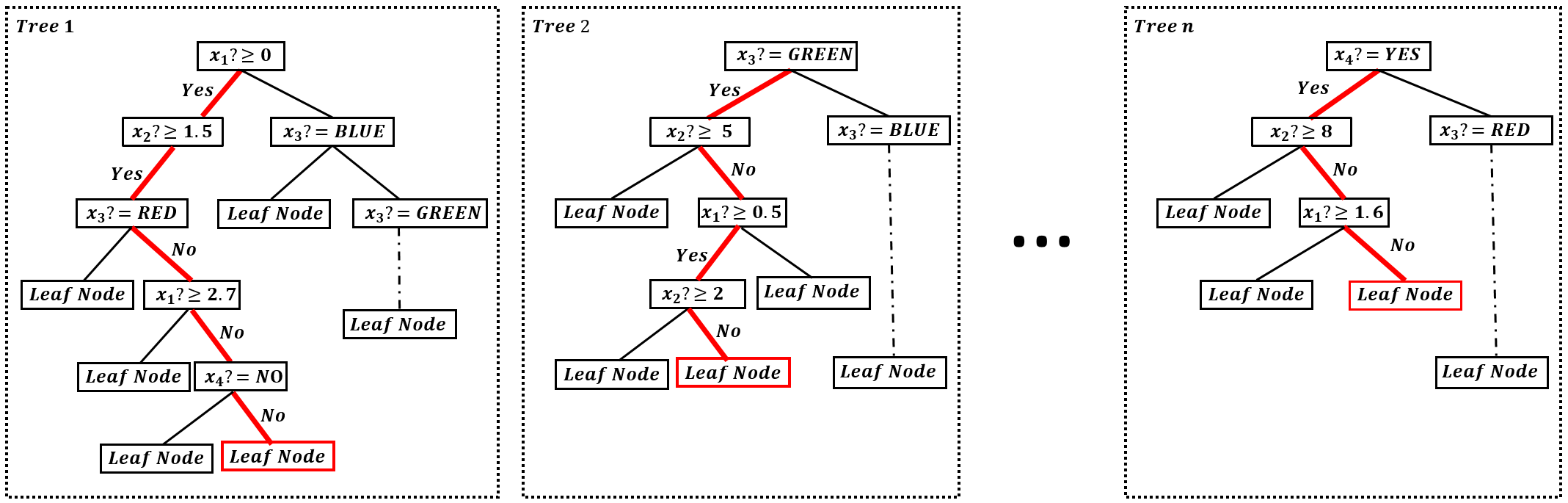}
   \caption{Traversing backward along decision paths }
   \label{fig:subregion}
\end{figure*}

Now suppose in the encoding step we have generated a forest as shown in Fig~\ref{fig:subregion}. Now, we only know the leaf nodes on which the instance $\bm{x}$ falling into, as shown in Fig~\ref{fig:subregion} as the red nodes, and wish to reconstruct $\bm{x}$.

Here, we propose an effective yet simple, possibly the simplest, strategy for backward reconstruction in forests.
First, each leaf node actually corresponds to a path coming from the root, we can identify the path based on the leaf node without uncertainty.

For example, in Fig~\ref{fig:subregion} the identified paths are highlighted in red color.
Second, each path corresponds to a symbolic rule; for example, the highlighted tree paths correspond to the following rule set,
where $RULE_i$ corresponds to the path of the $i$-th tree in the forest,
where $\neg$ denotes the negation of a judgment :

$RULE_1$: $(x_1 \geq 0)\wedge (x_2 \geq 1.5) \wedge \neg(x_3==RED) \wedge \neg(x_1 \geq 2.7) \wedge \neg(x_4==NO)$\\

$RULE_2$: $(x_3 == GREEN) \wedge \neg(x_2 \geq 5) \wedge (x_1 \geq  0.5) \wedge \neg(x_2 \geq 2)$  \\

... \\

$RULE_n: (x_4 == YES) \wedge \neg(x_2 \geq 8) \wedge \neg(x_1 \geq 1.6)$  

This rule set can be further adjusted into a more succinct form:

$RULE_1':   (2.7 \geq x_1 \geq 0) \wedge ( x_2 \geq 1.5) \wedge \neg(x_3==RED) \wedge (x_4==YES)$  \\

$RULE_2': (x_1 \geq 0.5) \wedge \neg(x_2 \geq 2) \wedge (x_3 == GREEN)$ \\

...\\

$RULE_n': \neg(x_1 \geq 1.6) \wedge \neg(x_2 \geq 8) \wedge  (x_4 == YES)$

Then, we can derive the Maximal-Compatible Rule (MCR). MCR is such a rule that each of its component coverage cannot be enlarged, otherwise incompatible issue will occur. For example,from the above rule set we can get the corresponding MCR:

$(1.6 \geq x_1 \geq 0.5) \wedge (2 \geq x_2  \geq 1.5) \wedge (x_3 == GREEN) \wedge (x_4 == YES)$

For each component of this MCR, such as  $(2 \geq x_2  \geq{ 1.5})$,
its coverage cannot be enlarged; for example, if it were enlarged to  $(3 \geq x_2\geq 1.5)$, it would have conflict with the condition in $\neg(x_2 \geq 2)$ in $RULE_2$. A more detailed description is shown in Algorithm~\ref{alg:MCR}.

It is very easy to prove the following theorem, and thus we omit the proof.

\begin{theorem}
The original sample must reside in the input region defined by the MCR.
\end{theorem}

Thus, after obtaining the MCR, we can reconstruct the original sample. For categorical attributes such as $x_3$ and $x_4$, the original sample must take these values in the MCR; for numerical attributes, such as $x_2$, we can take a representative value, such as the mean value in (2, 1.5). Thus, the reconstructed sample is ${\bm x}$ = [0.55, 1.75, GREEN, YES]. Note that for numerical value, we can have many alternative ways for the reconstruction, such as the median, max, min, or even calculate the histograms.

\begin{algorithm}[H]
\caption{Calculate MCR}
\label{alg:MCR}
\KwIn{A list $Rule\_List$ consists of $T$ rules defined by a forest}
\KwOut{$MCR$}

$MCR$ = initialize\_list()  \\
\For{$i$ in $range(T)$}
{\ 
$path\_rule$= $rule\_list[i]$\\
\For{$node\_rule$ in path\_rule.node\_rule\_list}
	{\ 
	j = node\_rule.attribute \\
	$MCR[j]$ = $intersect(MCR[j],node\_rule.bound)$  
	}
}
return MCR \\
\end{algorithm}

Given the above description, now we give a summary for conducting backward decoding of eForest. Concretely, given a trained forest with $T$ trees along with the forward encoding $x_{enc}$ in $R^{T}$ for a particular data, the backward decoding will first locate the individual leaf node via each element in $x_{enc}$, and then obtain $T$ decision rules for the corresponding decision paths accordingly. Then, by calculating the MCR, we can thus get a reconstruction from  $x_{enc}$ back to $x_{dec}$ in the input region. A concrete algorithm is shown in Algorithm~\ref{alg:Decoding}.

By enabling the eForest to conduct the forward encoding and backward decoding operations, autoencoding tasks can thus be realized. In addition, although beyond the scope of this paper, the eForest model might give some insight on a theoretical treatment for the representation learning ability for tree ensemble models, as well as helping to design new models for deep forest.

\begin{algorithm}[H]
\caption{Backward Decoding}\label{alg:Decoding}
\KwIn{$x_{enc}$, trained eforest $F$ with $T$ trees}
\KwOut{$x_{dec}$}
rule\_list = list() \\
\For{$i$ in range($T$)}
{\
path = forest.tree[i].get\_path($x_{enc}[i]$ ) \\
$path\_rule$ = calculate\_rules(path) \\
$path\_rule$ = simplify($path\_rule$)\\
rule\_list.append(path\_rule)  
}
 $MCR$ = calculate\_MCR(rule\_list) \\
$x_{dec}$ = sample($MCR$) \\
return  $x_{dec}$ \\
\end{algorithm}

\section{Experiments}
\subsection{Image Reconstruction}

We evaluate the performance of eForest in both supervised and unsupervised setting. In this implementation, we take Random Forest \citep{Breiman2001} to construct the supervised forest, whereas take the completely-random forest \citep{Zhou:gcForest17} as the routine for the unsupervised forest. Notice that other decision tree ensemble construction methods can also be used for this purpose. Concretely, for supervised eForest, each non-terminal node randomly select $\sqrt{d}$ attributes in the input space and pick the best possible split for information gain; for unsupervised eForest, each non-terminal node randomly pick one attributes and make a random split.
In our experiments we simply grow the trees to pure leaf, or terminate when there are only two instances in a node. We evaluate eForest containing 500 trees or 1,000 trees, denoted by $eForest_{500}$ and $eForest_{1000}$ respectively. Note that $eForest_N$ will re-represent the input instance as a $N$-dimensional vector.

Since auto-encoders especially DNN-based auto-encoders are mainly designed for image tasks, in this section we run some experiments on image data first. We use the MNIST dataset \citep{LeCun:Bottou1998}, which consists of 60,000 gray scale 28$\times$28 images (784 dimensional vector per sample) for training and 10,000 for testing. We also use CIFAR-10 dataset \citep{krizhevsky2009learning}, which is a more complex dataset consists of 50,000 colored 32$\times$32 images (therefore each image is in $R^{1024}$ per channel) for training and 10,000 colored images for testing. For colored images, the eForest process each channel separately for memory saving.

\begin{table}[h]
\centering
\caption{Performance comparison (measured by MSE). The subscript $s$ and $u$ denote supervised and unsupervised, respectively.}\label{table:ae}

\centering
\begin{tabular}{|c|c|c|}
\hline
                                 & MNIST                 & CIFAR-10 \\ \hline
$MLP_{1}$                  & 266.85                & 1284.98        \\ \hline
$MLP_{2}$                   & 163.97               & 1226.52       \\ \hline
CNN                              &  768.02                &   865.63       \\ \hline
$eForest^{s}_{500}$       & 1386.96               & 1623.93      \\ \hline
$eForest^{s}_{1000}$       & 701.99                & 567.64      \\ \hline
$eForest^{u}_{500}$        & 27.39                 & 579.337  \\ \hline
$eForest^{u}_{1000}$       & \textbf{6.86}         & \textbf{153.68}  \\ \hline

\end{tabular}

%
\end{table}

MLP based AutoEncoders (MLP-AEs) and a convolutional neural network based auto-encoder (CNN-AE) are used for comparison. For MLP-AEs, we follow the suggestions in \citep{bengio2007greedy} and use two architectures, with 500-dimensional and 1000-dimensional inner representation, respectively. Concretely, the MLP-AE $MLP_{1}$ for MNIST is $(input-1024-500-1024-output)$ and the $MLP_{2}$ for MNIST is $(input-2048-1000-2048-output)$. Likewise, the MLP-AE $MLP_{1}$ for CIFAR-10 is $(input-4096-1024-500-1024-4096-output)$ and the $MLP_{2}$ for CIFAR-10 is $(input-4096-2048-1000-2048-4096-output)$. For CNN-AE, we follow the implementations in the Keras documentation \footnote{https://blog.keras.io/building-autoencoders-in-keras.html} with the following architecture: It consisting of a conv-layers with 16 (3 $\times$ 3) kernels followed by 2 conv-layers with 8 (3 $\times$ 3) kernels, and each conv-layer has a 2 $\times$ 2 maxpooling layer followed. The decoder we used  has same structure as encoder except using up-sampling layer instead of pooling layers (for mapping the data back to its original input space). ReLUs are used for activations and logloss is used as training objective. During training, dropout is set to be 0.25 per layer.

Experimental results are summarized in Table \ref{table:ae}. For DNN auto-encoders, cross validation are used for hyper-parameter tuning; for eForest, we just take the min value of the interval defined by the corresponding MCR as indicated in the last sampling step of decoding.

\begin{figure}[!h]%
\centering 
\subfloat[Reconstructed samples on CIFAR-10] {\includegraphics[scale=0.418]{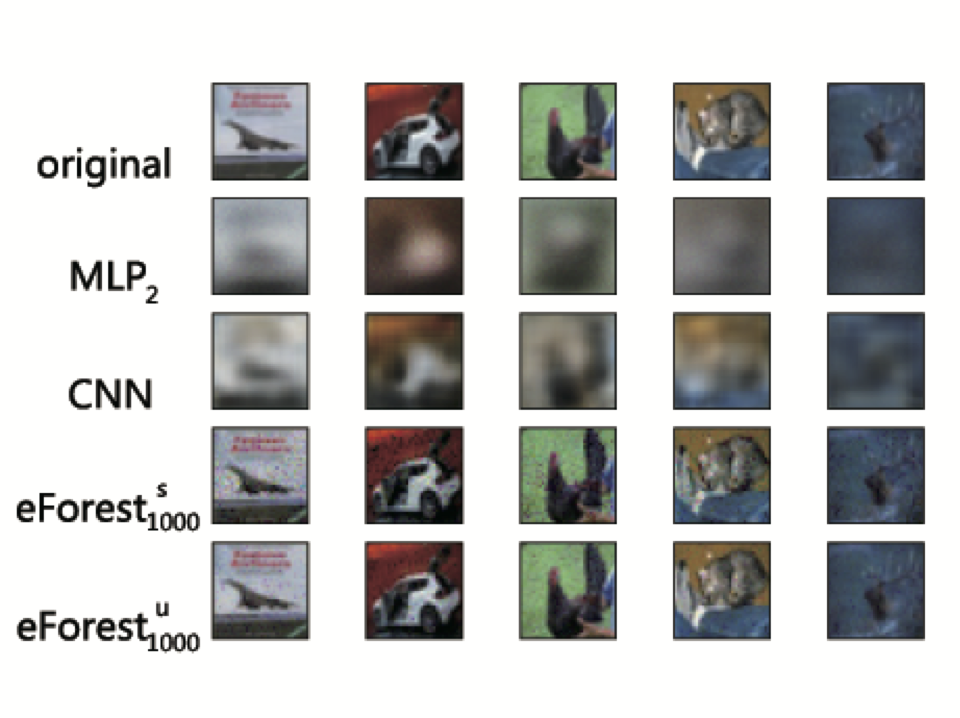}}
\subfloat[Reconstructed samples on MNIST]{\includegraphics[scale=0.418]{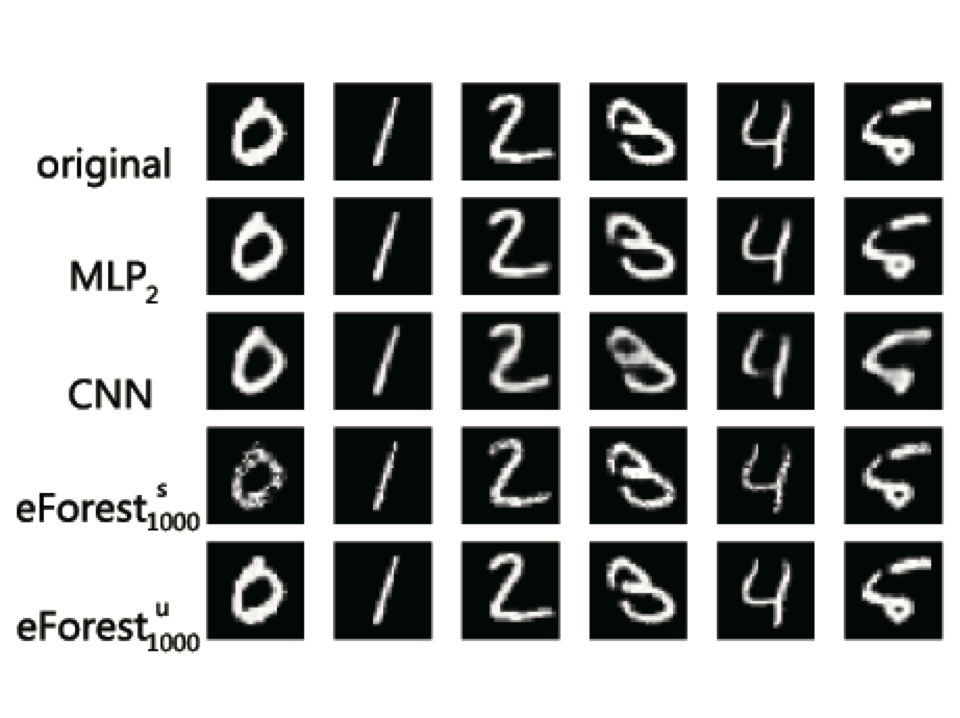}}
\caption{The original test samples (first rows) and the reconstructed samples}
\label{fig:recon}
\end{figure}

It can be seen that eForest achieves the best performance. Some reconstructed samples on the test set are shown in Figure~\ref{fig:recon}. This result looks sad for CNN based auto-encoders on CIFAR-10 dataset, as we are using the architecture recommended for image auto-encoders by Keras documentation and have carefully tuned the other hyper-parameters via cross-validation. We believe that the DNN autoencoders can get improved performance by some further tuning; nevertheless, the eForest auto-encoder works well without careful parameter tuning.

It is worth noting that the unsupervised eForest had a better performance compared with the supervised eForest, given the same number of trees. Note that each decision tree path corresponds to a rule, whereas a longer rule will define a tighter MCR. We conjecture that a tighter MCR might lead to a more accurate reconstruction. Therefore for a forest with longer tree depth may have a better performance. For example, we measured the maximum depth as well as the average depth for all trees on MNIST dataset, as summarized in Tabel  ~\ref{table:length}. Experimental results give positive supports, as shown in Table~ \ref{table:length}. An unsupervised eForest indeed has a longer average depth.

\begin{table}[!h]
\centering
\caption{Length of tree depth on MNIST}\label{table:length}
\centering
\begin{tabular}{|c|c|c|}
\hline
                                & Max depth                 & Ave. depth \\ \hline
$eForest^{s}_{500}$       & 48               & 34.82      \\ \hline
$eForest^{s}_{1000}$      & 48                & 34.79      \\ \hline
$eForest^{u}_{500}$        & 93                & 70.87  \\ \hline
$eForest^{u}_{1000}$       & 101         & 70.07  \\ \hline
\end{tabular}
\end{table}

\subsection{Text Reconstruction}
In addition to image tasks, other tasks may also require auto-encoders. Thus, we study the performance of eForest for text reconstruction. Note that the DNN auto-encoders are mainly designed for images, and if to be applied to texts, some additional mechanism such as word2vec embedding\citep{mikolov2013distributed} is required for pre-processing. Here, in our experiments, we want to study the performance of doing auto-encoding directly on text data.

Concretely, we used the IMDB dataset \citep{Maas:Daly:Pham2011} which contains 25,000 documents for training and 25,000 documents for testing. Each document was stored as a 5,000 dimensional vector via tf/idf transformation. We used exactly the \textbf{same} configuration of eForests for image data. Cosine distance is used for evaluation metric, which is the standard metric for measuring the similarities between documents represented by tf/idf vectors. The lower the cosine distance, the better. The results are summarized in Table ~\ref{table:imdb}.

\begin{table}[!h]
\centering
\caption{Text reconstruction}\label{table:imdb}

\begin{tabular}{|c|c|c|}
\hline
                                 & Cosine Distance               \\ \hline
$eForest^{s}_{500}$       & 0.1132                     \\ \hline
$eForest^{s}_{1000}$       & 0.0676                    \\ \hline
$eForest^{u}_{500}$        & 0.0070                   \\ \hline
$eForest^{u}_{1000}$       & \textbf{0.0023}         \\ \hline
\end{tabular}
\end{table}

It should be highlighted that CNN based auto-encoders can not be applied on this kind of input data at all and MLP based auto-encoders is barely useful. After extensive cross-validation for parameter search, the best structure for MLP we could obtained is $(Input-4096-2048-1024-2048-4096-Output)$, with the performance of 0.512, more than two hundred times worse than eForest.

From the above results, we showed that eForest can also be applied on text data with high performance. In addition, notice that by using only $10 \%$ of the bits of representation (eForest of 500 trees trained unsupervisedly), eForest can already reconstruct the original input with high accuracy. This is a promising result which can be further utilized for data compression.

\subsection{Computation Efficiency}
As a common advantage for tree ensemble models, eForest is also inherently apt for parallel implementation. We implement eForest on a single KNL-7250 (belongs to Intel XEON Phi many-core product family), and achieved a 67.7 speedup for training 1,000 trees in an unsupervised setting, compared with a serial implementation. For a comparison, we trained the corresponding MLPs and CNN-AEs with the same configurations as in the previous sections on one Titan-X GPU and the results for training cost as well as testing per sample cost are summarized in the Table ~\ref{tab:HPC}.

\begin{table}[!h]
\centering
\caption{Time cost (in seconds). Decoding time is measured in sample per seconds.}\label{tab:HPC}
\begin{tabular}{|l|l|l|l|l|}
\hline
\multirow{2}{*}{Models} & \multicolumn{2}{l|}{MNIST} & \multicolumn{2}{l|}{CIFAR-10} \\ \cline{2-5}
                  & Train     & Decode    & Train       & Decode      \\ \hline
$eForest_{1000}$   & 22.174       & 0.725       & 66.201        & 4.296         \\ \hline
$MLP_{2}$      & 274.757      & 0.003       & 1606.89       & 0.004         \\ \hline
CNN               & 214.901      & 0.021       & 253.57        & 0.021         \\ \hline
\end{tabular}
\end{table}
 
From the above results, eForest is more than 100 times fast when training, but is slower during encoding time than DNN based auto-encoders. We hope that the decoding can be speedup by some more optimization in the future.

\subsection{Damage Tolerable}

There are cases when the model is partially damaged due to a various reasons such as memory or disk failure.
For a partially damaged model is still able to function in such cases is one characteristic towards model robustness.
The eForest approach for auto-encoding is one such model by its nature since we could still estimate the MCR when facing only a subset of trees in the forest.

\begin{figure}[!h]%
\centering
\subfloat[MNIST] {\includegraphics[scale=0.4518]{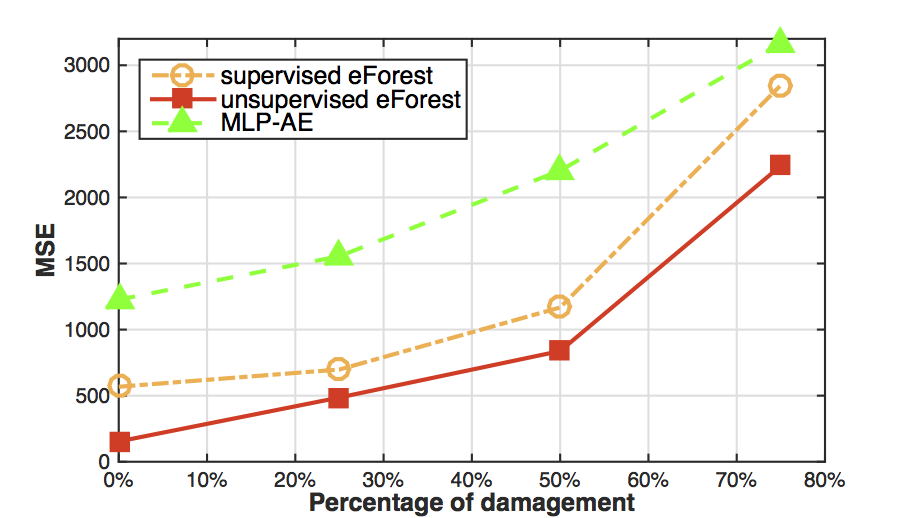}}
\subfloat[CIFAR-10]{\includegraphics[scale=0.4518]{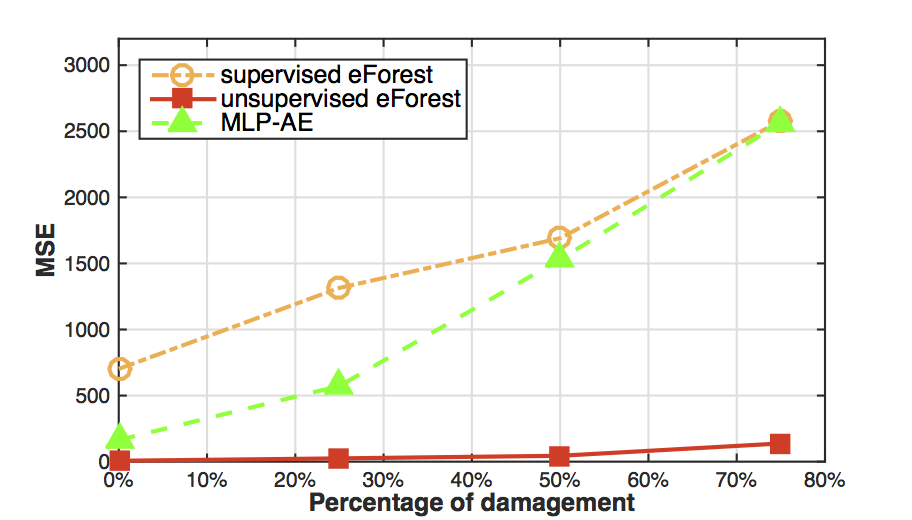}}
\caption{Performance when model is partially damaged.}
\label{fig:sec2_part}
\end{figure}

In this section, we test the damage tolerable empirically on CIFAR-10 and MNIST datasets. Concretely, during testing time, we randomly drop 25\%, 50\% and 75\% of the trees and measure the reconstruction error based on the pattern recovered using only the remaining trees. For a comparison, we also randomly turned off 25\%, 50\% and 75\% of the neurons in the $MLP_{2}$ with structure exactly the same as in the previous section. The performance results are illustrated in Figure~\ref{fig:sec2_part}.

Form the above result, the eForest approach is more damage tolerable than a MLP-AE, and the unsupervised eForest is the most damage tolerable model among others.

\subsection{Model Reuse for eForest}

In an open environment, the test data for encoding/decoding may belong to a different distribution with the training data. In this section, we test the ability for model reuse and the goal here is to train a model in one dataset and reuse it in another dataset without any modifications or re-training. The ability for model reuse in this context is an important property for future machine learning developments \citep{learnware}.

\begin{figure}[!h]%
\centering
\subfloat[Reconstructed omniglot samples by models trained from mnist.] {\includegraphics[scale=0.328]{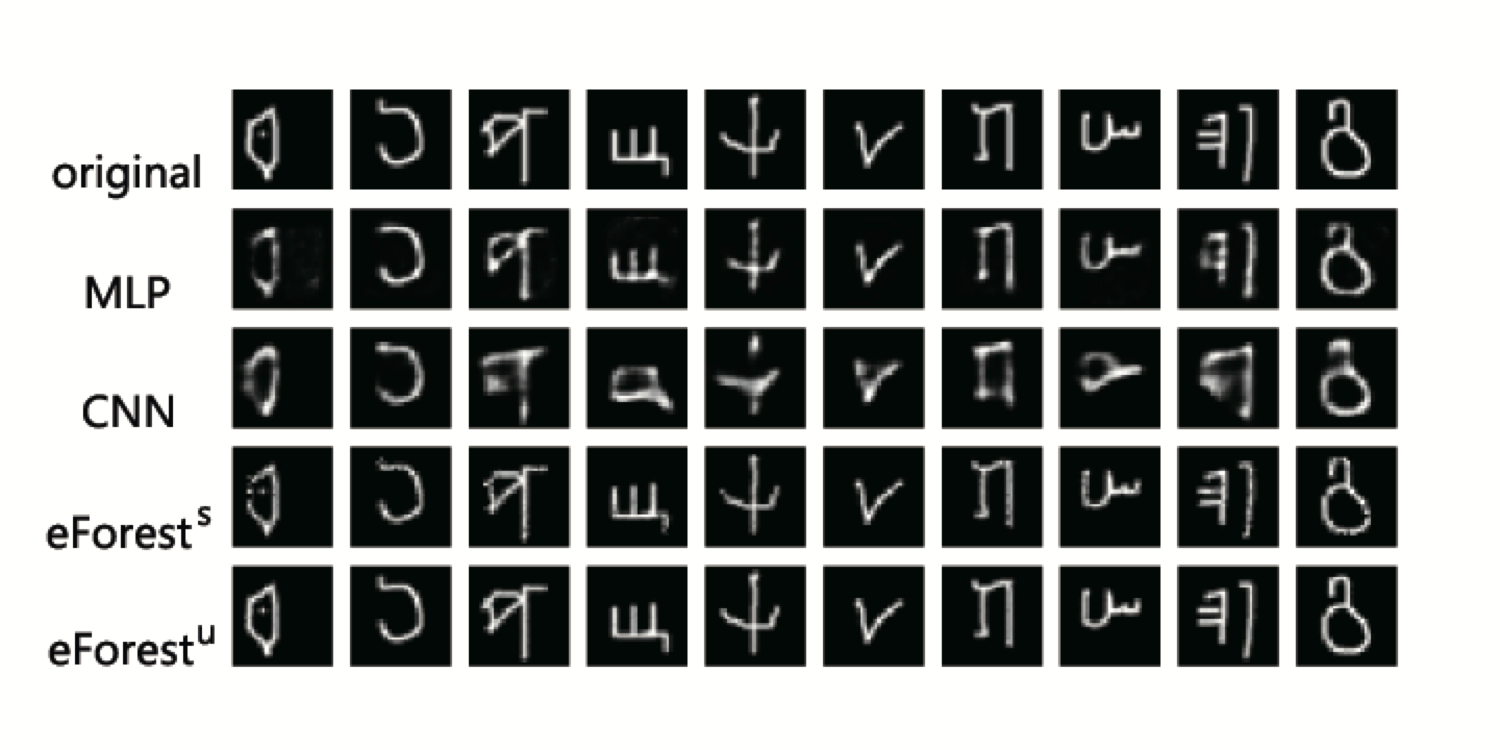}}

\subfloat[Reconstructed mnist samples by models trained from cifar.]{\includegraphics[scale=0.328]{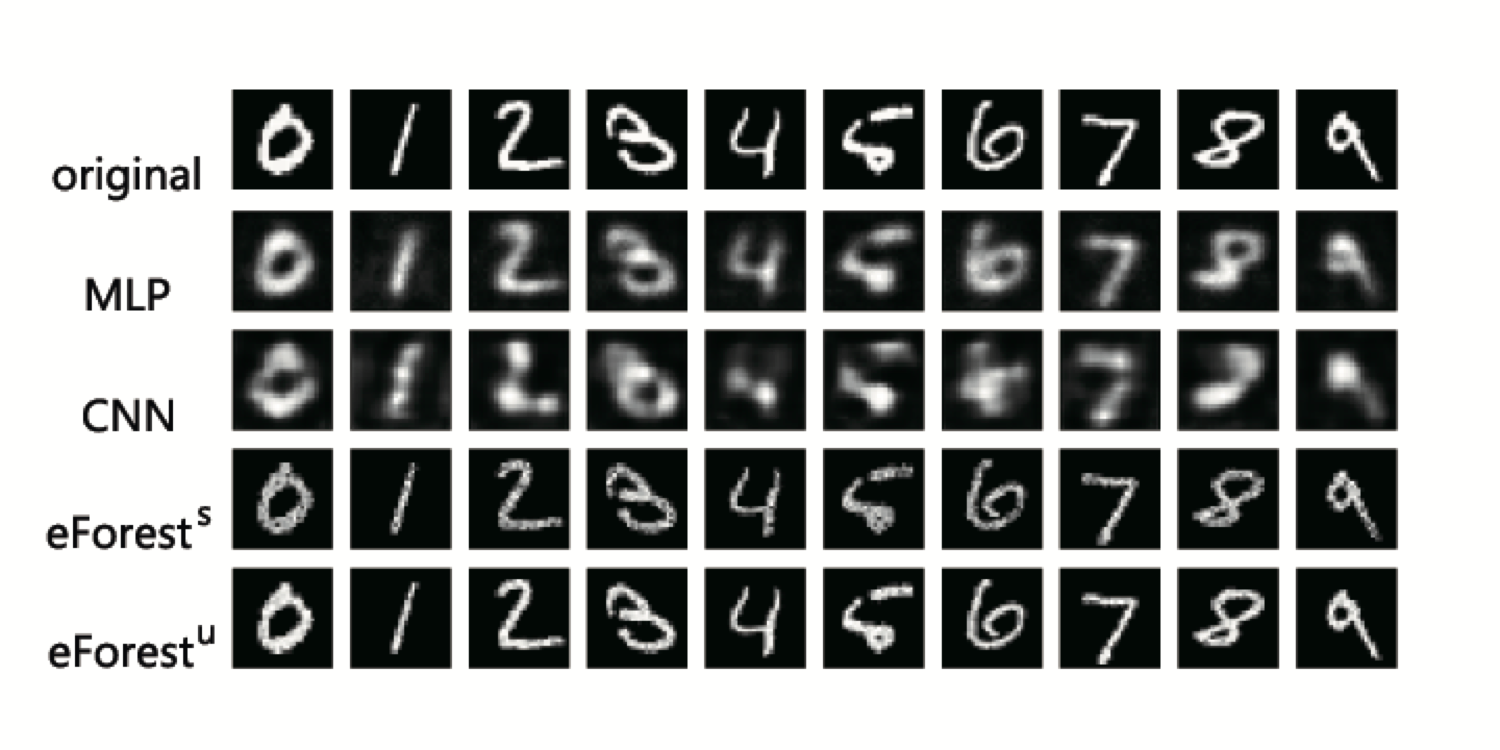}}
\caption{The original samples(first rows) and the ones reconstructed by different AEs, where $eForest_{s/u}$ correspond to supervised/unsupervised setting, respectively.}
\label{fig:mnist2omni}
\end{figure}

Concretely, we evaluate the ability for model reuse as follows. We trained an unsupervised and an supervised eForest on CIFAR-10 dataset (converted and rescaled to 28$\times$28 gray scale data), each consisting of 1,000 trees , and then use the exact models to encoding/decoding data from the MNIST test dataset. Likewise, we also trained eForests consists of 1,000 trees on MNIST dataset, and directly test the encoding/decoding performance on the Omniglot datasets \citep{omini}. For a fair comparison, we trained a CNN-Autoencoder and MLP-Autoencoder on the same dataset without fine-tuning. The architecture for MLP/CNN-AEs and the training procedures are the same in the previous sections accordingly. MSE is used for performance evaluation.

\begin{table}[h]
\centering
\caption{Performance comparison for model reuse (measured by MSE).}\label{table:autoencoder1}
\begin{tabular}{|l|l|l|l|}
\hline
\multirow{2}{*}{Model} & cifar train & \multirow{2}{*}{Model} & mnist train  \\
                       & mnist test  &                        & omniglot test \\ \hline
$MLP_{2}$               & 1898.76          & $MLP_{2}$     &     596.24      \\ \hline
CNN                & 2657.69          & CNN      &     1280.60      \\ \hline
$eForest^{s}$         & 652.38         & $eForest^{s}$  &  270.54         \\ \hline
$eForest^{u}$          & \textbf{90.43} & $eForest^{u}$  & \textbf{12.80}          \\ \hline
\end{tabular}
\label{tab:cifar2mnist}
\end{table}

Some randomly picked reconstructed samples are presented in Fig.~\ref{fig:mnist2omni}, and the numerical evaluation on the whole test set is presented in Table ~\ref{tab:cifar2mnist}. It can be inferred that eForests has out-performed the DNN approach by a factor more than 100.  Specifically, for an eForest trained on CIFAR-10 can perform a better encoding/decoding task on MNIST dataset, and these two dataset are quite different. It showed the generalization ability in terms of model reuse for eForest.

\section{Conclusion}
In this paper, we propose the EncoderForest (abbrv. eForest),  the first tree ensemble based auto-encoder model, by devising an effective procedure for enabling forests to reconstruct the original pattern by utilizing the Maximal-Compatible Rule (MCR) defined by decision paths of the trees. Experiments demonstrate its good performance in terms of accuracy and speed, as well as the ability of damage tolerance and model reusability. In particular, on text data, by using only $10\%$ of the input bits, the model is still able to reconstruct the original data with high accuracy. Another advantage of eForest lies in the fact that it can be applied to symbolic attributes or mixed attributes directly, without transforming the symbolic attributes to numerical ones, especially when considering that the transforming procedure generally either lose information or introduce additional bias.

Note that supervised and unsupervised eForest are actually the two ingredients utilized simultaneously in each level of the deep forest constructed by gcForst. This work might offer some additional understanding of gcForst\citep{Zhou:gcForest17}. Constructing a deep eForest model is also an interesting future issue.

\bibliographystyle{elsarticle-harv}
\bibliography{nips17}
\end{document}